\documentclass{article}
\usepackage{ijcai21}

\usepackage{times}
\usepackage{soul}
\usepackage{url}
\usepackage[hidelinks]{hyperref}
\usepackage[utf8]{inputenc}
\usepackage[small]{caption}
\usepackage{graphicx}
\usepackage{amsmath}
\usepackage{amsthm}
\usepackage{booktabs}
\usepackage{algorithm}
\usepackage{algorithmic}
\usepackage{graphicx}
\usepackage{verbatim }
\usepackage{subcaption}
\usepackage{tikz}
\usepackage{bm}
 \usetikzlibrary{positioning}
\usepackage{multirow}
\title{A Streaming End-to-End Framework For Spoken Language Understanding}

\author{
Nihal Potdar\footnote{The work was done during a coop internship in Noah's Ark Lab.}$^1$\and
Anderson R. Avila$^2$\and
Chao Xing$^2$\and
Dong Wang$^{3}$\and
Yiran Cao$^{*1}$\and
Xiao Chen$^2$
\affiliations
$^1$University of Waterloo\\
$^2$Huawei Noah's Ark Lab\\
$^3$Tsinghua University
\emails
\{npotdar, y267cao\}@uwaterloo.ca, \{anderson.avila, xingchao.ml, chen.xiao2\}@huawei.com,
wangdong99@mails.tsinghua.edu.cn
}

\begin{document}

\maketitle

\begin{abstract}

End-to-end spoken language understanding (SLU) has recently attracted increasing interest. 
Compared to the conventional tandem-based approach that combines speech recognition and 
language understanding as separate modules, the new approach extracts users' intentions directly from the speech signals, resulting in joint optimization and low latency. 
Such an approach, however, is typically designed to process one intention at a time,
which leads users to take multiple rounds to fulfill their
requirements while interacting with a dialogue system. In this paper, we propose 
a streaming end-to-end framework that can process multiple intentions in an online and
incremental way. The backbone of our framework is 
a unidirectional RNN trained with the connectionist temporal classification (CTC) criterion. 
By this design, an intention can be identified when sufficient evidence has been accumulated, and 
multiple intentions can be identified sequentially. 
We evaluate our solution on the Fluent Speech Commands (FSC) 
dataset and the intent detection accuracy is about 97 \% on all multi-intent settings.
This result is comparable to the performance of the state-of-the-art non-streaming models, but
is achieved in an online and incremental way. We also employ our model to a keyword spotting task using the Google Speech Commands dataset
and the results are also highly promising. 

\end{abstract}

\vspace{6pt}
\section{Introduction}

\begin{figure}
  \centering
  \includegraphics[scale=0.42]{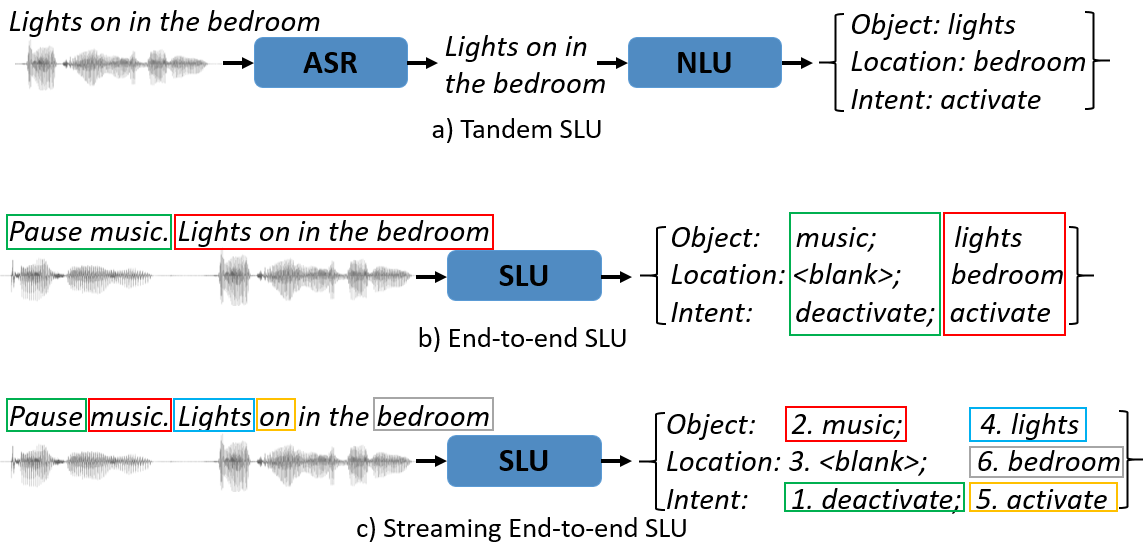}
  \caption{Intent classification based on (a) a tandem SLU system, comprised of automatic speech recognition and natural language understanding, capable of processing only a single intent at a time; (b) an end-to-end SLU that classifies multiple intents from the speech signal, enabling multi-turn dialogue; (c) a streaming end-to-end SLU which incrementally predicts intents while processing incoming chunks of the speech signal.}
  \label{fig:conventional_slu}
\end{figure}
\begin{figure*}
\centering
  \includegraphics[width=0.8\linewidth]{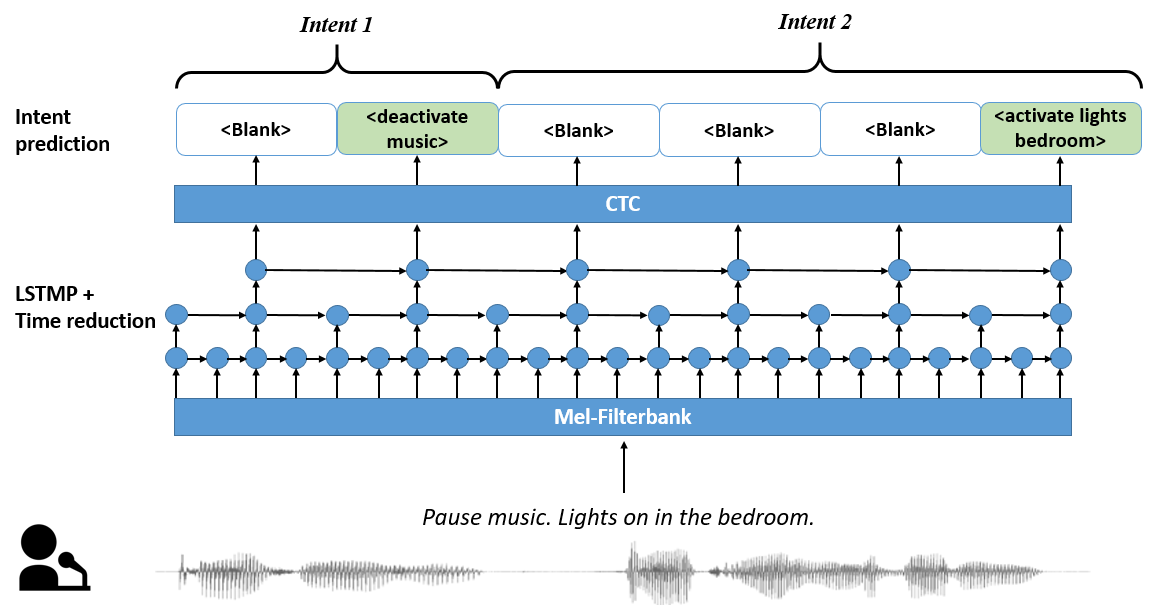}
  \caption{The proposed RNN-CTC based model enables the output to be updated in real-time as sufficient evidence is accumulated from the speech signal, hence allowing the identification of multiple intentions.}
  \label{fig:model_architecture}
\end{figure*}

Spoken language understanding (SLU) aims at extracting structured semantic representations, such as intents and slots, from the speech signal \cite{mhiri2020low}. For instance, a speech waveform carrying out the sentence ``Lights on in the bedroom" can be represented as \{intent:`activate lights' , location: `bedroom'\} (see Figure~\ref{fig:conventional_slu}). Given the proliferation of voice user interfaces (VUIs) in smartphones and other gadgets, these representations are crucial to provide voice control capabilities to these devices, enabling speech as a primary mode of human-computer interaction (HCI). SLU systems are particularly useful for hands-free conversational interactions, such as while driving a car, cooking or accessing a smart home system.

Tandem SLU solutions rely on the text transcription generated by an automatic speech recognition (ASR) module. The ASR output is then processed by a natural language understanding (NLU) system, responsible for interpreting the transcript content \cite{qian2019spoken,shivakumar2019incremental}. This approach is represented in Figure~\ref{fig:conventional_slu} (a). 
A notable disadvantage of this approach relates to the fact that the ASR and the NLU systems are optimized separately, therefore not globally optimal for the target SLU task. Moreover, 
the NLU is often optimized on clean text while the ASR output usually contains errors. This causes severe error accumulation and amplification, aggravating the overall SLU performance \cite{mhiri2020low}.
Although joint training approaches have been proposed, e.g., for the language models ~\cite{liu2016joint}, the tandem nature prevents a full joint optimization.

Recently, end-to-end SLU has been proposed to tackle this problem. As shown in Figure~\ref{fig:conventional_slu} (b), 
the new approach extracts users’ intentions directly from the speech signals, resulting in joint optimization~\cite{haghani2018audio,palogiannidi2020end,lugosch2019speech,chen2018spoken,mhiri2020low}.
Despite the promising progress, almost all the existing end-to-end SLU methods need to wait at least until all the speech segments are filled or the end of the speech utterance is detected by other models before starting the intent prediction, thus suffering from 
significant latency~\cite{mhiri2020low,haghani2018audio}. This latency will lead to delayed feedback during human-computer interactions, which often seriously impacts users' experience~\cite{porcheron2018voice}. 
For example, in \cite{kohrs2016delays}, the authors conduct a systematic investigation based on functional Magnetic Resonance Imaging (fMRI) showing 
that delayed feedback (noticeable around 327.2-ms $\pm 89.7$) can affect the behaviour, the psychophysiology as well as the neural activity of the user. In \cite{schoenenberg2014you}, it is shown that users in far-end conversation are much less attentive if the delay is more than 400 ms.
To reduce the latency, some recent researches are proposed to process the speech signal segment by segment~\cite{mhiri2020low}, 
however the intent prediction is still based on the complete utterance. 

Another deficiency of the existing end-to-end SLU methods is that they are short in dealing with multiple intentions. Although users can be asked to speak simply and show one intention,
people tend to speak more complex utterances with multiple intents in a single round of interaction. For example, people may say `I want to book a flight and if possible, please reserve a receiving taxi for me'.
For most existing end-to-end approach, the system can only deal with one intent, perhaps the more significant one `reserve taxi'. This 
will invoke additional interactions to deal with the missing intent. 

In this paper, we propose a fully streaming end-to-end framework for SLU, as shown in Fig.~\ref{fig:conventional_slu} (c). 
The meaning of `streaming' is two-fold: (1) the intent is identified `online', which means one intention 
is identified whenever sufficient evidence has been accumulated, without the need to wait until the end of the utterance, which significantly reduce the latency; (2) the intent 
is identified `incremental', which means multiple intentions can be identified sequentially. To the best of the authors' knowledge, this is the first streaming end-to-end architecture 
for the SLU task.

To achieve that, a unidirectional recurrent neural network (RNN), trained with the connectionist temporal classification (CTC) method, is proposed. 
The framework is evaluated on three underline tasks. First, our model is assessed on single-intent classification and experiments are conducted on the Fluent Speech Commands corpus (FSC) \cite{lugosch2019speech}. 
Second, to evaluate the proposed framework on multi-intent settings, we generate versions of the FSC dataset containing speech samples with multiple intents. Details about the data generation is given in Section~\ref{sec:experiments}. 
Third, on keyword spotting, where the task refers to detecting short speech commands \cite{warden2018speech}. 
For that, we use two versions of the Google Speech Commands (GSC) dataset \cite{warden2018speech}. 

The remainder of this paper is organized as follows. In Section \ref{sec:related_work}, we review the related work on end-to-end SLU and approaches to
reduce latency. 
Section \ref{sec:model_arch} presents the proposed framework. 
Section \ref{sec:experiments} describes our experimental setup and Section~\ref{sec:experimental_results} 
discusses our results. Section \ref{sec:conclusion} gives the conclusion and future works.

\begin{figure*}
\centering
  \includegraphics[width=0.8\linewidth]{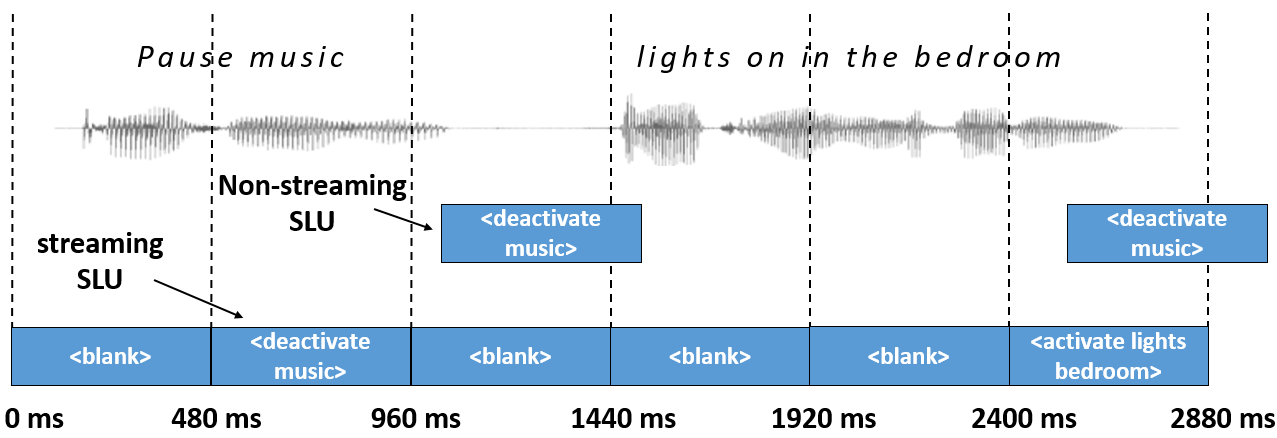}
  \caption{Illustration of non-streaming SLU and streaming SLU. Non-streaming SLU collects the speech frames of the entire utterance and predicts an intent at the end. In contrast, streaming SLU predicts the intent whenever the evidence for the intent is sufficient. Therefore, streaming SLU may predict an intent before the utterance is end, and can predict multiple intentions in a sequential way.}
  \label{fig:ctc}
\end{figure*}

\vspace{6pt}
\section{Related Work}
\label{sec:related_work}

Our work is based on the end-to-end SLU approach. Compared to the tandem ASR + SLU architecture, the end-to-end approach identifies users' intent from speech 
directly, hence avoiding the notorious error propagation through individual modules. A key problem of the end-to-end approach is the limited training data, due to
the great variety of the speech signal when expressing an intent. Researchers have made extensive investigation and achieved promising progress. 
For example, in \cite{haghani2018audio}, several end-to-end SLU encoder-decoder solutions are investigated. 

In \cite{palogiannidi2020end}, an end-to-end RNN-based SLU solution is proposed. Several architectures for predicting intents conditioned to the slot filling task are introduced. The authors demonstrate that data augmentation is required in order to attain the state-of-the-art performance while training deep architectures for end-to-end SLU. Another recent study introduces the Fluent Speech Command dataset and also presents a pre-training methodology for end-to-end SLU models. Their approach is based on adopting ASR targets, such as words and phonemes, that are used to pre-train initial layers of their final model. These classifiers, once trained, are discarded and the embeddings from the pre-trained layers are used as features for the SLU task. The authors show improved performance on large and small SLU training sets with the proposed pre-training strategy \cite{lugosch2019speech}. A similar approach is found in \cite{chen2018spoken}. The authors also propose to fine-tune the front layers of an end-to-end CNN-RNN model that learns to predict graphemes. This pre-trained acoustic model is optimized with the CTC loss,
which allows the model to learn the alignments between input and output sequences. This acoustic model is then combined with a semantic model to predict intents. 
Note that the CTC loss is calculated for the ASR module, not for the SLU model as in our work.

Treatments of latency have been investigated in the tandem ASR+NLU architecture, in principle based on segment-level ASR output. 
For example, \cite{shivakumar2019incremental} propose a recurrent neural network (RNN) that processes ASR output of 
arbitrary length and provides intent predictions in real-time. 
Similarly, \cite{liu2016joint} uses an RNN to perform online intent detection and slot filling jointly as ASR output arrives. 
For the end-to-end system, \cite{mhiri2020low} propose a speech-to-intent model 
based on a global max-pooling layer that allows for processing speech sequentially while receiving upcoming segments.
Although the authors are able to achieve low latency in their results, the solution is not online, as
the intent prediction still requires the entire sentence to be processed completely. 

Finally, incremental multiple intention detection has been studied with the ASR+NLU architecture~\cite{shivakumar2019incremental}, based on 
segmental ASR output and online and incremental NLU. For the end-to-end architecture, no extensive research has been found.

Compared to the aforementioned research, our proposal is a complete streaming end-to-end SLU framework, capable of incrementally predicting multiple intents 
from incoming chunks of the speech signal. 
Next, we shall give the reader details on our proposed solution.

\vspace{6pt}

\section{Proposed Framework}
\label{sec:model_arch}
We start this section by presenting the main ideas and challenges behind modeling a streaming end-to-end SLU solution. Next, we introduce the proposed framework. We show how CTC optimization can help our framework to perform multi-intent classification without requiring prior alignment knowledge between the input and output sequences. We finish this section by giving details on how to train our proposed framework with a two-stage pre-training approach.
\vspace{6pt}

\subsection{Principle}
\noindent \paragraph{End-to-end SLU.} An ASR-free end-to-end SLU model aims at classifying a speech utterance, represented by a sequence of acoustic features $X$, as a single semantic label $L$ (or intent class), hence bypassing the need of an explicit combination of an ASR and an NLU modules~\cite{qian2019spoken}. This is achieved by maximizing the following posterior probability,

\begin{equation}
    L^* = \arg\max_{L} P(L|X, \theta)
\end{equation}

\noindent where $\theta$ is the parameters of the end-to-end neural network model. Compared to the tandem ASR + NLU architecture, the end-to-end approach solves the error propagation problem, but usually requires large amounts of data to deal with the high variation in speech signal when expressing the same intent. Methods including data augmentation, pre-training and multi-task learning have been widely used~\cite{palogiannidi2020end,lugosch2019speech,chen2018spoken}.

\noindent \paragraph{End-to-end streaming SLU.} The above end-to-end approach predicts the intent with the entire utterance, and can only predict one intent for a single utterance. This approach is non-streaming. For a streaming approach, however, the model is expected to perform a sequence-to-sequence prediction, which predicts an intent when the evidence is sufficiently accumulated, and multiple intents can be predicted incrementally. The reader can refer to Figure~\ref{fig:ctc} as illustration of the difference between the non-streaming and streaming approaches. Compared to the non-streaming
case, the primary difficulty for the streaming model is how to accumulate evidence for each intent so that appropriate intents can be `fired' when the evidence is sufficient. In the model that will be presented shortly, we employ CTC training as the major tool to solve this problem.

\vspace{6pt}
\subsection{Proposed Architecture}
We formulate the streaming end-to-end SLU as a sequence-to-sequence task, where our goal is to predict the sequence of intents $L$ by given an input sequence of acoustic features $X$. The proposed architecture is shown in Fig.~\ref{fig:model_architecture}. The main structure of the model is a unidirectional RNN network, where the units are long short-term memory with projection (LSTMP). A potential problem for SLU is that the lengths of the speech input and the intent labels are significantly different. The same problem arises in ASR, but for SLU it is more evident as an intention often corresponds to a much longer speech segment than a word. In order to condense the semantic information and reduce the length discrepancy, time reduction~\cite{chan2015listen} was employed to reduce the frame rate for each layer. 

\vspace{6pt}
\subsection{CTC Criterion}
\label{subsec:ctc}

The key problem of the proposed streaming end-to-end architecture is how to fire intentions \emph{automatically} when evidence is sufficient. We resort to the Connectionist Temporal Classification (CTC) training. The CTC training was first motivated to train RNNs of unsegmented data \cite{graves2006connectionist}. Previous to CTC, training RNNs required prior segmentation of the input sequence. With such segmentation, each feature was labelled with an appropriate target and the RNN was trained to predict the target step by step, via
the back-propagation through time (BPTT). With CTC, all the potential alignments between the input and output sequences are considered during the training, by inserting blank tokens ($<$blank$>$) between any output labels (including the beginning and end of the sequence) to ensure the input and output sequences have the same length. This approach has been successfully employed in various tasks including speech recognition~\cite{miao2016empirical} and machine translation~\cite{libovicky2018end}.

Formally, the conditional probability of a single alignment, $\alpha$, is the product of the probabilities of observing $\alpha_t$ at time $t$ and can be represented as

\begin{equation}
    P(\alpha|X, \theta) = \prod_{t=1}^{T} P(\alpha_t | X, \theta)
\end{equation}

\noindent where $\alpha_t$ represents the given label (including the blank token) at time $t$, according to the alignment $\alpha$. 
Because $P(\alpha|X)$ defines mutually exclusive alignment, the conditional probability for the target sequence $Y$ given the input sequence $X$ amounts to the sum of the probabilities 
associated with all the alignments:

\begin{equation}
    P(L|X, \theta) = \sum_{\alpha \in \mathcal{B}_{X,Y}} \prod_{t=1}^T P(\alpha_t|X, \theta)
\end{equation}

\noindent where $\mathcal{B}_{X,L}$ is the set of all valid alignments. The CTC loss is then defined as

\begin{equation}
    L_{CTC}(X,L) = - \log \sum_{\alpha \in \mathcal{B}_{X,L}} \prod_{t=1}^T P(\alpha_t|X, \theta)
\end{equation}

\noindent This model can be efficiently trained following a forward-back algorithm~\cite{graves2006connectionist}. A favorable feature of a CTC model is that during decoding, true target labels are output only the evidence for them are sufficiently accumulated, otherwise the model will output the blank token~\cite{sak2015fast}. Once a label has been output, its evidence is reset and a new round of evidence accumulation is started. This feature perfectly fits the request of the streaming SLU, where we expect the model to fire an intent if sufficient evidence from the speech signal has been accumulated, and once it is fired, the evidence should be accumulated again. By this setting, multiple intents can be sequentially identified, no matter how long the speech signal is.

\vspace{6pt}
\subsection{Two-stage Pre-training}
\label{subsec:two-stage}
Although CTC can potentially solve the streaming SLU task, some training skills are still required in order to tackle the intrinsic data sparsity problem (as in non-streaming models). For that, we designed a two-stage pre-training scheme as described below. Our experiments show that the pre-training methods are crucial for training a high-quality streaming end-to-end model.

\vspace{6pt}
\noindent \paragraph{ASR pre-training.} In the first stage, we initialize the feature extraction layers with a pre-trained ASR model. Only the first layer was used for this purpose. After training, the weights were kept frozen and used to provide embeddings for the next LSTM layers. The upper LSTM layers use these embeddings (or features) to make predictions. This procedure aims at leveraging the large speech corpora and the fine-grained word labels in ASR. For that, we utilize the 100 hours of speech signals in the Librispeech dataset \cite{panayotov2015librispeech}. Our pre-trained model is based on classifying character-level targets, employing the CTC loss. Using characters instead of words as the training target simplifies the training task~\cite{lugosch2019speech}. In this study, we only reuse the first layer of the ASR model for the SLU model, which ideally plays the role of robust feature extraction. Different from the approach investigated in \cite{kuo2020end} and \cite{lugosch2019speech}, where the weights of the pre-trained ASR model were fine-tuned while training the SLU model, the layer borrowed from ASR model in our study remains frozen.

\vspace{6pt}
\noindent \paragraph{Cross-entropy pre-training.} In the second stage, we pre-train the SLU model (the first layer fixed) with a cross-entropy (CE) loss, before taking the CTC training. For simplicity, we compute the CE loss at the last time step only and back-propagate to the beginning of the sequence. After this pre-training, the model is fine-tuned by using the CTC loss.

\vspace{6pt}
\section{Experimental Setup}
\label{sec:experiments}

We evaluate the proposed model in three tasks: single-intent classification, multi-intent prediction and spoken keyword spotting. 
In this section, we firstly present the datasets used and then describe the experimental settings. 


\vspace{6pt}
\subsection{Datasets}

Four datasets were involved in our experiments, all audio clips were (re)sampled at 16 kHz.

\vspace{6pt}
\begin{itemize}

\item All of our pretrained ASR models were trained on the LibriSpeech corpus, which was introduced in \cite{panayotov2015librispeech}. It was derived from audiobooks and contains roughly 1000 hours of speech. We selected 100 hours of clean speech to train the ASR model. There are couple of reasons for that. First, we are proposing a compact model with relatively reduced number of parameters. Second, the pre-trained ASR is attained by optimizing the weights in the first layer, which means even fewer parameters to be optimized. Therefore, we considered 100 hours (3x more than the in-domain FSC data) enough to train the first layer. Note that this layer is kept frozen while training the rest of the network. Empirically, we found no benefit in unfreezing the first layer while training our model with in-domain data. Similar findings have been found for the same dataset in previous work \cite{lugosch2019speech}.

\item The Fluent Speech Commands (FSC) dataset \cite{lugosch2019speech} was used to train and evaluate our SLU model for intent classification. It comprised single-channel audio clips collected using crowd sourcing. Participants were requested to speak random phrases for each intent twice. The dataset contained about 19 hours of speech, providing a total of 30,043 utterances spoken by 97 different speakers. Each utterance contains three slots: action, object, and location. We considered a single intent as the combination of all the slots (action, object and location), resulting 31 intents in total.
The whole dataset was split to three subsets: the training set (FSC-Tr) contained 14.7 hours of data, totalling 23,132 utterances from 77 speakers; the validation set (FSC-Val) and test set (FSC-Tst) comprised 1.9 and 2.4 hours of speech, leading to 3,118 utterances from 10 speakers and 3,793 utterances from other 10 speakers, respectively.
The validation set was used to select the model, while the test set was used to evaluate the SLU performance.

\item To train and test the multi-intent prediction model, we generated three additional multi-intent datasets using the utterances from the FSC dataset. 
The first dataset, namely FSC-M2, was attained by concatenating two utterances from the same speaker into a single sentence. We have made sure that all the possible intent combinations were evenly distributed in the dataset. 
The training part of FSC-M2 (FSC-M2-Tr) contained 57,923 utterances (74.27 hours) selected from the FSC training data, and the test part (FSC-M2-Tst) 
contained 8,538 utterances (11.80 hours) from the FSC test data. Following a similar procedure, we constructed a three-intent dataset (FSC-M3)
by augmenting the utterances in FSC-M2 with a new utterance from FSC. Finally, we combined roughly 40 \% of random samples from FSC, FSC-M2, and FSC-M3 to build a large multi-intent set (FSC-MM),
where the data splitting for training and test was the same as the original datasets.
For model selection, we also constructed validation sets for each multi-intent condition.

\item We also used the Google Speech Commands (GSC) dataset \cite{warden2018speech} to verify the capability of our model in spotting short key words. 
The provider of the dataset published two versions. Version 1 comprised 64,727 utterances from 1,881 speakers, a total of 30 words. We split the dataset into training, validation and test sets, containing 14.08, 1.88, 1.89 hours of speech signals respectively. Version 2 was larger, consisting of 105,829 utterances of 35 words, recorded by 2,618 speakers. The training, validation and test sets contain 23.30, 3.02, 2.74 hours of speech, respectively. 

\end{itemize}

\vspace{6pt}
\subsection{Experimental Settings}
\label{sec:features}
In this work, audio signals are sampled at 16 kHz. 
The Kaldi toolkit is used to extract 80-dimensional log Mel-Filterbank 
features with 3-dimensional pitch, totalling 83 dimensions. To extract the Mel features, the audio signal is processed in frames of 320 samples (i.e., 20-ms window length), 
with a step size of 160 samples (that is, 10-ms hop-size). Global Cepstral Mean and Variance Normalization (CMVN) are 
applied in order to mitigate the mismatch between training and testing data.

The model was trained using the ADAM optimizer \cite{loshchilov2017decoupled}, with the initial learning rate set to $0.0001$. 
Dropout probability was set to 0.1 and the parameter for weight decay was set to 0.2.
For the ASR pre-training, the ASR model was trained 100 epochs; for the CE pre-training, the model was
trained for 10 epochs with the CE criterion.





\vspace{6pt}
\section{Experimental Results}
\label{sec:experimental_results}

We present the results on three tasks: single-intent classification, multi-intent prediction and spoken keyword spotting. 
For the ablation study, 
note that we present the results with and without the two pre-training techniques. Results are all reported on test sets.


\begin{table}
\centering
\scalebox{0.8}{
\begin{tabular}{lcc}
\toprule
Model & Streaming & Accuracy\\
\midrule
RNN+Pre-training \cite{lugosch2019speech} & no & 98.8 \\
CNN+Segment pooling \cite{mhiri2020low} & no & 97.8 \\
CNN+GRU(SotA) \cite{tian2020improving} & no & 99.1 \\
\midrule
RNN+CTC & yes & 12.66 \\
+ ASR Pre-training & yes & 15.83 \\
+ CE Pre-training & yes & 98.90 \\
\bottomrule
\end{tabular}}
\caption{Experimental results on FSC for single-intent classification. Performance is reported in terms of accuracy (\%).}
\label{tab:res_single}
\end{table}

\vspace{6pt}
\subsection{Experiment I: Single-intent Classification}
\label{subsec:exp1}

The first experiment is single-intent classification, conducted with the FSC dataset. 
The model has three LSTMP layers, each with $1024$ hidden states followed by 
a projection layer with 320 dimension. As described in Section~\ref{subsec:two-stage}, the first layer is based on a pre-trained ASR model where the current 
frame is stacked with 7 frames to the left. Frame-skipping is employed, with only 1/3 frames need to be processed, leading to
faster computation. 
The next two LSTMP layers involve a time reduction operation with the reduction factor $4$. 
This results in a 480-ms input context for each output time step. 
The output of the third LSTMP layer is fed to two fully connected layers and the final output corresponds to the 31 intents. 

Table~\ref{tab:res_single} presents the results for this task. 
For comparison, three recent end-to-end SLU solutions are adopted as benchmarks:
the RNN model with pre-training presented by~\cite{lugosch2019speech}, the CNN model with segment pooling presented
by~\cite{mhiri2020low}, and the CNN trained with Reptile presented by~\cite{tian2020improving}. The CNN+Reptile 
approach represents the state-of-the-art (SotA) performance on this task. We 
emphasize that all of the three models are non-streaming and so cannot perform online and incremental predictions. 
Nevertheless, their results set good reference for our streaming model.

The results in Table~\ref{tab:res_single} demonstrated that our proposed streaming model is highly effective: it achieves performance as high as 98.90 \% with 
the two-stage pre-training. This result outperforms the first two benchmarks, and is very close to the SotA model \cite{tian2020improving}, while our
approach is streaming. 

Focusing on the pre-training methods, it can be observed that both the ASR pre-training and the CE pre-training 
improve the performance, though the contribution of the CE pre-training is much more significant. This is not surprising as the 
supervision from the target (intent) is very sparse and usually appears at the end of an utterance. The CE pre-training offers a strong bias on the more reasonable alignments for the CTC training.

An interesting observation is that our streaming model usually predicts the intent
before the end of the utterance. In most cases, the spotting position is at the 2/3 length of the 
utterance. This implies that the model has learned to accumulate the evidence and can make the 
prediction decision if the evidence has been strong enough, rather than waiting for the end of the utterance. This feature is highly desirable. It enables a truly online SLU, and it is a major advantage that the CTC model possesses.

\vspace{6pt}
\subsection{Experiment II: Multi-intent Prediction}

In the second experiment, we evaluate our model's ability to predict multiple intents.
For that, we adopt our architecture based on the two-stage pre-training approach, 
which provides the best results in the first experiment. 
We first use the model trained on the single-intent 
corpus (FSC-Tr) to predict multiple intents, and then retrain multi-intent models using 
multi-intent corpora. Note that
the benchmark models in the previous experiment cannot predict multiple intents, so we did not
report them here.

\begin{table}
\centering
\scalebox{0.75}{
\begin{tabular}{clcccc}
\toprule
 && \multicolumn{4}{c}{\text{Trained}} \\
\cmidrule{3-6}
 &&FSC-Tr &FSC-M2-Tr&FSC-M3-Tr&FSC-MM-Tr\\
\midrule
\multirow{3}{*}{\smash{\rotatebox[origin=c]{90}{\text{Tested}}}}&FSC-Tst & \textbf{98.90} & 98.77 & 97.96 & \textbf{98.90}  \\
&FSC-M2-Tst & 56.90  & \textbf{97.93} & 96.90  & 97.89   \\
&FSC-M3-Tst& 26.73  & 97.15& 97.01  & \textbf{97.28}  \\
\bottomrule
\end{tabular}}
\caption{Experimental results for the multi-intent prediction. Performance is reported in terms of accuracy (\%).}
\label{tab:res_multi}
\end{table}

The performance is presented in Table \ref{tab:res_multi},
where each row represents a test condition and each column represents a training condition. Note that for the multi-intent
test, only if all the intents involved in the utterance are identified, the result is recognized as correct. Therefore, 
the multi-intent prediction is much more challenging than the single-intent classification.

\begin{table}
\centering
\scalebox{0.86}{
\begin{tabular}{lcc}
\toprule
Model & V1 & V2 \\
\midrule
Att-RNN \cite{de2018neural} & 95.6 & 96.9 \\
TC-ResNet \cite{choi2019temporal} &  - & 96.6 \\
SincConv+DSConv \cite{s2020small} & 96.6 & 97.4 \\
TENET12+MTConv \cite{li2020smallfootprint} &  96.84 & - \\
\midrule
CTC & 62.64 & 62.98 \\
+ Pre-trained ASR & 64.16 & 63.59 \\
+ Pre-trained CE & 96.35 & 97.10 \\
\bottomrule
\end{tabular}}
\caption{Experimental results on the keyword spotting task. Performance is reported in terms of accuracy (\%) on the GSC dataset V1 and V2.}
\label{tab:res_keyword}
\end{table}

Table \ref{tab:res_multi} shows that, when trained with single-intent data and tested on two-intent and three-intent utterances, 
the performance of our streaming model is significantly degraded, specially in the three-intent case. 
This is mainly because there is no prior knowledge about the intent boundaries involved in the training data. 
Nonetheless, the single-intent model can still predict multiple intents in some cases, especially in the two-intent setting.
This is surprising and demonstrated that the CTC streaming model is rather powerful and has learned how to 
accumulate evidence for intents and reset it when an intent has been detected. 

\begin{figure*}
\centering
\begin{subfigure}{.5\textwidth}
  \centering
  \includegraphics[width=.9\linewidth,height=6cm]{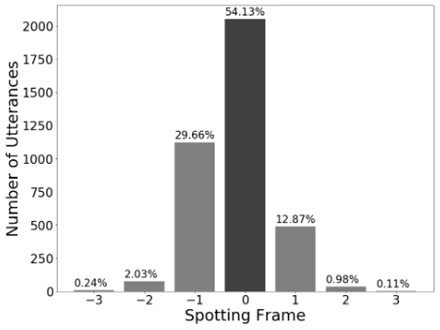}
  \caption{}
  \label{fig:sub1}
\end{subfigure}%
\begin{subfigure}{.5\textwidth}
  \centering
  \includegraphics[width=0.9\linewidth, height=6cm]{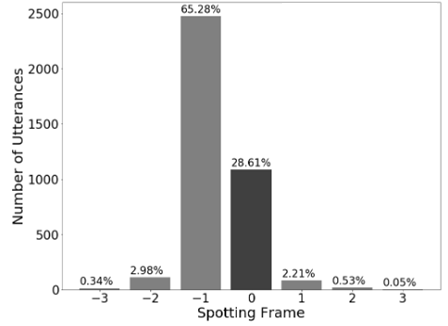}
  \caption{}
  \label{fig:sub2}
\end{subfigure}
\label{fig:stats}
\caption{Relative spotting positions compared to ground truth, with (a) model trained and evaluated on single intent experiment and (b) model trained on multi-intents setting and evaluated first intent. The x-axis represents the relative distance between the spotting position and the ground truth position. Negative values refer to early spotting and positive values refer to late spotting.}
\end{figure*}

When multi-intent data are employed to train the model, the performance on the two-intent and three-intent conditions 
is significantly improved. 
Surprisingly, the accuracy of the multi-intent prediction is almost the same as the single-intent prediction, 
although the former is much harder. This implies that by learning the multi-intent samples, 
the model has well learned how to reset the evidence quickly, so that it can prepare for the next intent prediction in time. 
Remarkably, the model seems to learn this knowledge with two-intent data only and this knowledge can be generalized to 
conditions with more intentions. For example, on the three-intent test, 
the accuracy was increased from 26\% to 97\% when we trained the model with two-intent data.




\vspace{6pt}
\subsection{Experiment III: Keyword Spotting}

In this last experiment, we customize the streaming end-to-end architecture and apply it to a 
keyword spotting task. This model consists of two LSTMP layers. 
The first LSTMP layer involves $1024$ hidden states, followed by a projection layer with $320$ units. 
This layer is initialized by ASR pre-training as in the intention prediction experiment. 
The second LSTMP layer is similar, but time reduction is applied, so that the time scope
is enlarged by $4$ times. 
The output from the second LSTMP layer is fed into two fully connected layers, and the output corresponds to the keywords for detecting.

Two models were trained using the training set of GSC V1 and V2 respectively, and were tested on the test set of GSC V1 and V2 individually. For the test on each dataset, 12 keywords were selected. 
The performance in terms of detection accuracy is presented in Table ~\ref{tab:res_keyword}. For comparison, we reported the results of 
four benchmarks that represent the state-of-the-art research on this dataset. 
Results show that on both the two test sets, our model based on two-stage pre-training offers competitive performance when compared to the benchmark results. Meanwhile, our model is streaming and holds the great advantage on spotting multiple 
keywords online and incrementally. 

\vspace{6pt}
\section{Analysis for Early Spotting}

\begin{figure}
\centering
\includegraphics[width=.99\linewidth, height=5cm]{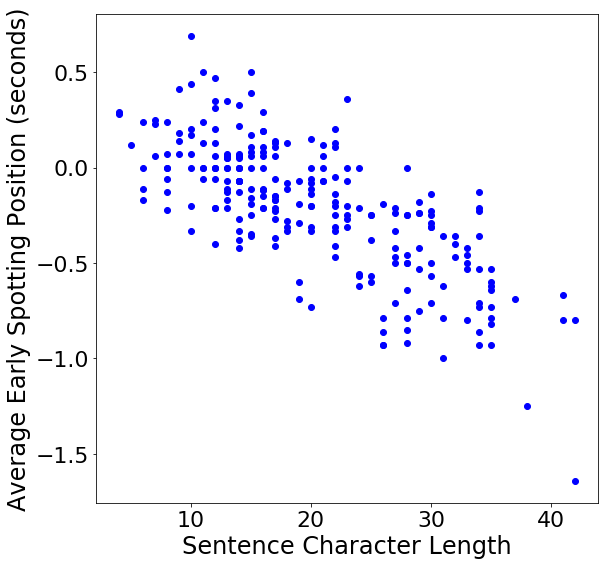}

\caption{The relation between the early-spotting behavior and the sentence length in characters. The x-axis represents the sentence length, and the y-axis represents the relative spotting position. }
\label{fig:ling}
\end{figure}

A key advantage of our streaming SLU system is that intents or keywords can be determined whenever evidence is sufficient. In this section, we present some statistical analysis of such early-spotting behavior based on the intent detection. The first analysis investigates the statistics of relative position of the spottings, i.e., the distance between the spotting positions and ground truth positions. Note that we treat the last non-silence frame corresponding to an intention as its ground truth position, which can be determined by Voice Activity Detection (VAD). Figure~\ref{fig:sub1} shows the relative positions of the spottings in the intent detection experiments, where a negative distance denotes an early spotting, and a positive distance denotes a late spotting. It can be found that in the single-intent scenario, more than 30\% of the spottings occur earlier than the ground truth positions, and this proportion is 68\% in the multi-intent scenario. The second analysis investigates the relation between the early-spotting behavior and the length of the sentence. The results are shown in Figure~\ref{fig:ling}, where the x-axis represents the sentence length in characteres, and the y-axis represents the relative spotting position. It can be found that long sentences are more likely to be early spotted than short sentences.

\vspace{6pt}
\section{Conclusion}
\label{sec:conclusion}

In this paper, we propose an end-to-end streaming SLU framework, which can predict multiple intentions in an online and incremental way. 
Our model is based on a unidirectional LSTM architecture, optimized with the alignment-free CTC loss and pre-trained with the cross-entropy criterion.  We test the model on the intent prediction task. Results show that our model achieves comparable performance to start-of-the-art non-streaming models for single-intent classification. For multi-intent prediction, which is more challenging and impossible for the existing non-streaming models, our approach almost maintains the accuracy of single-intent classification.  The model was also tested on a keyword spotting task, and the performance is competitive compared to four benchmarks. As future work, we aim to explore streaming intent + slot classification, and integrate local attention mechanism into our streaming solution.
\vspace{6pt}


\bibliographystyle{named}
\bibliography{ijcai21_v1}

\end{document}